# Noise-robust classification with hypergraph neural network

**Nguyen Trinh Vu Dang[1], Loc Tran[2], Linh Tran[3]**
[1,3]Ho Chi Minh City, University of Technology (HCMUT), VNU-HCM, Ho Chi Minh City, Vietnam
[2]Laboratoire CHArt EA4004 EPHE-PSL University, Vietnam

*Corresponding Author:*

Linh Tran
Department of Electronics
Ho Chi Minh City University of Technology (HCMUT), VNU-HCM
268 Ly Thuong Kiet Street, District 10, Ho Chi Minh City, Vietnam
Email: linhtran@hcmut.edu.vn

## 1. INTRODUCTION

During the last decade, the deep convolution neural network can be considered the current state of the art method for various classification tasks such as image recognition [1], speech recognition [2], to name a few. Recently, to deal with irregular data structures, data scientists have gained many interests in graph convolution neural network method such as [3]. In this method, the pairwise relationships between objects (samples) are used. In the other words, in this graph data structure, the edge of the graph can connect only two vertices.

To overcome the information loss due to only considering the "pairwise relationship between objects" of graph data structure [4, 5] have recently proposed the hypergraph neural network approach. In this hypergraph data structure, an edge (hyperedge) can connect more than two vertices. In the other words, the hy- peredge is the subset of the set of vertices of the hypergraph. Recently, this hypergraph neural network method has just been employed to solve classification tasks [4, 5] and outperforms the graph neural network and can be considered the current state of the art method of semi-supervised learning approach. However, this method has also not been utilized to solve the noisy label learning problem.

Inspired from the idea combining the pagerank algorithm with the graph convolution neural network in [6], in this paper, we propose the novel version of hypergraph neural network method combining the classic hypergraph based semi-supervised learning method [7, 8] with the hypergraph neural network method [4, 5]. In the other words, we combine the propagation scheme utilizing the hypergraph model with the hypergraph neural network which is the current state of the art method of semi-supervised learning approach. We find out that this proposed combination of the propagation scheme and the hypergraph neural network method significantly improves the accuracy of the hypergraph neural network method alone even when the noise presents in the labels.

In this paper, our contributions are three-folds:
a) In order to reduce the runtime constructing the graphs and the hypergraphs from the image datasets, we apply the dimensional reduction technique PCA to the image datasets.
b) Propose the novel version of hypergraph neural network method combining the classic hypergraph based semi-supervised learning method with the hypergraph neural network method.
c) Compare the accuracy performance measures of the classic graph based semi-supervised learning problem, the classic hypergraph based semi-supervised learning problem, the graph neural network method, the hypergraph neural network method, and our proposed hypergraph neural network method when we apply these five methods to solve the noisy label learning problem.

We will organize the paper as follows: Section 2 will discuss the related work. Section 3 will introduce the novel version of hypergraph neural network method. Section 4 will describe the datasets and present the experimental results. Section 5 will conclude this paper and the future direction of researches will be discussed.

## 2. RELATED WORK
Learning with label noise gains many interests since label noise may lead to many undesirable concerns such as the decrease in learning performance. The current studies associated to this problem can be assembled into three main groups [9]:
a) Robust model approach
b) Data filtering approach
c) Inherently noise-tolerant learning approach

### 2.1. Robust model approach
This approach empirically studies the robustness property of various classical classification algorithms such as Naïve Bayes probabilistic classifier, C4.5 decision tree, the SMO support vector machine [9], to name a few. Experimental results show that the Naïve Bayes probabilistic classifier and the random forest ensemble classifier are the most robust classical classification systems against noise label [9]. However, the weakness of these classical classification is very clear. First, this approach is inactive. Instead of refining and changing the classical classification algorithms, they only explore the robustness prop- erty of commonly used classification systems. Second, the most robust classification system is effective only when the percentage of label noise is minor. The performance of the most robust classification system drops significantly when the percentage of label noise is huge.

Recently, the deep neural networks (i.e., the modern classification systems) have been established and well developed. For example, the deep convolution neural network can be considering the current state of the art and the best classification system for image recognition problem [1]. However, [10-12] showed that a deep neural network with huge enough capacity can memorize the training set labels even when they are randomly made. Hence, they are mostly vulnerable to the label noise. Similar to classical classification systems, label noise can cause overfitting and significantly drop the deep neural networks' performance.

However, [12] observed that when the learning rate is high, deep neural networks may preserve quite extraordinary accuracy. In the other words, the influence of the label noise is not important. This observation was employed in [12] to preserve an approximation of the labels using the running average of deep neural network's forecasts with a high learning rate. These approximations then can be utilized as the control (or supervision) signals to train the deep neural network. Inspired by the work of [10-12] and [4-6, 13], in this paper, we will develop the graph and hypergraph convolution neural network methods and apply these methods to solve the "noisy label learning" problem (using the image datasets such as MNIST, USPS, and FASHION MNIST). To the best of our knowledge, this work has not been investigated and developed.

### 2.2. Data filtering approach
In this approach, the samples with noisy labels are distinguished and fixed before the training process. The ruined labels can be merely eliminated or relabeled at the very beginning. One idea is to employ the forecasts from the classification system (for e.g. the SVM system) to detect mislabeled samples. The class of these methods is called the classification filtering system. However, filtering all the samples that are misclassified by the classification filtering system is too inflexible and risky since the classification filtering system learned from data with noisy labels might not always be accurate.

*k-nn* classification systems is related to another class of methods. k-nn classification systems are shown that they are very vulnerable to label noise. Hence some *k-nn* methods target at improving or changing the rules of *k-nn* algorithms to recognize noisy labels and then eliminate or relabel the associated samples. However, the methods associated to this class are heuristics and might not be effective for samples that are close to the classification boundary. Moreover, the choice of k (i.e. the number of neighbors) might affect the performance of these methods considerably.

Some data filtering methods depend on some thresholds. These methods compute the score for each sample by using some measures and eliminate the samples that are above the definite threshold. Obviously, these methods are similar to outlier or anomaly or abnormal detection methods which are very hard to solve. Moreover, it's very difficult to differentiate the accurate exemptions from the mislabeled samples. Last but not least, for the data filtering approach, this approach tends to eliminate a large amount of samples, which may contain key information for classification. Easily, we can also recognize that we can employ the forecasts from deep neural networks to detect mislabeled samples.

### 2.3. Inherently noise-tolerant learning approach
In this approach, there are two ways to attack the "noisy label learning problem":
a) First, this way will model the noisy label before or during the training phase to take the noisy label into attention. This model holds the information of the noise and can be inserted into the classification system [14-18]. However, in order to build the model for noisy label, these methods require supplementary parameters, that might increase the time complexity and the model complexity. We know that high model complexity occasionally leads to over-fitting.
b) Second, this way proposes the robust loss function for the noise-tolerant model. For example, [19] explored the robustness of various loss functions such as mean squared loss, mean absolute loss, and cross entropy loss. In [20], combined the benefits of the mean absolute loss and the cross entropy loss to attain the improved loss function.

## 3. HYPERGRAPH NEURAL NETWORK
### 3.1. Problem formulation
In this paper, we would like to solve the "noisy label learning" problem [10, 21]. This problem can also be called the "label distribution learning" problem [22], to name a few. In this problem, let $X_{train}=\{x_1, x_2, \ldots, x_l\}$ be the training set, where $x_i \in m$. $x_i$ can also be called the feature vector *i* or instance *i* or sample *i* of the training set with $1 \leq i \leq l$. Let $Y_L=\{y_1, y_2, \ldots, y_C\}$ be the complete set of labels where *C* is the number of classes in the dataset. For each sample $x_i$, there is a label distribution $d = \{d_{x_i}^{y_1}, d_{x_i}^{y_2}, \ldots, d_{x_i}^{y_c}\} \in R^C$. Please note that $d_{x_i}^{y_c}$ is the probability that the sample $x_i$ belongs to the class *c*. From the above definition, we know that $0 \leq d_{x_i}^{y_c} \leq 1$ and $\sum_c d_{x_i}^{y_c} = 1$.

The objective of the noisy label learning problem is to learn a mapping function $g: x \to d$ between the sample *x* and its corresponding label distribution function *d*. In the other words, the goal of noisy label learning is to learn the conditional probability mass function $p(y|x)$, where $y \in Y_L, x \in X_{train}$. Assume that $p(y|x)$ is the parametric model $p(y|x, \theta)$, where $\theta$ is the parameter vector. Given the training set $X_{train}$, we need to find (i.e., solve for) $\theta$ that can generate the distribution similar to *d* [23].

### 3.2. Preliminary notation and definitions
Given a hypergraph *G=(V, E)*, where *V* is the set of vertices and *E* is the set of hyper-edges. Each hyper-edge $e \in E$ is the subset of *V*. Please note that the cardinality of *e* is greater than or equal two. In the other words, $|e| \geq 2$, for every $e \in E$. Let *w(e)* be the weight of the hyper-edge *e*. Then *W* will be the $R^{|E|*|E|}$ diagonal matrix containing the weights of all hyper-edges in its diagonal entries. The incidence matrix *H* of *G* is a $R^{|V|*|E|}$ matrix that can be defined as follows:

$$h(v,e) = \begin{cases} 1 \text{ if vertex } v \text{ belongs to hyperedge } e \\ 0 \text{ otherwise} \end{cases} \quad (1)$$

The example of the hypergraphs is illustrated in Figure 1.

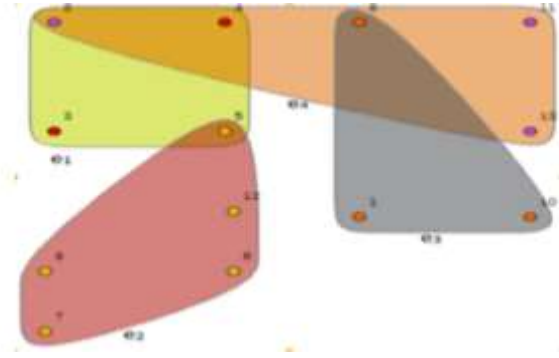

Figure 1. Hypergraph examples with 13 vertices and 4 hyperedges

From the above definition, we can define the degree of vertex *v* and the degree of hyper-edge *e* as follows:

$$d(v) = \sum_{e \in E} w(e) * h(v,e) \tag{2}$$

$$d(e) = \sum_{v \in V} h(v,e) \tag{3}$$

Let $D_v$ and $D_e$ be two diagonal matrices containing the degrees of vertices and the degrees of hyper-edges in their diagonal entries respectively. Please note that $D_v$ is the $R^{|V|*|V|}$ matrix and $D_e$ is the $R^{|E|*|E|}$ matrix. From the above definitions, [7, 8] define the symmetric normalized hypergraph Laplacian as follows:

$$L_{sym} = I - D_v^{-\frac{1}{2}} H W D_e^{-1} H^T D_v^{-\frac{1}{2}} \tag{4}$$

Moreover, [7, 8] define the random walk hypergraph Laplacian as follows:

$$L_{rw} = I - D_v^{-1} H W D_e^{-1} H^T \tag{5}$$

Please note that the two terms $D_v^{-\frac{1}{2}} H W D_e^{-1} H^T D_v^{-\frac{1}{2}}$ and $D_v^{-1} H W D_e^{-1} H^T$ in the symmetric normalized hypergraph Laplacian and the random walk hypergraph Laplacian respectively will be used in our proposed hypergraph neural network method.

### 3.3. Hypergraph based semi-supvervised learning problem

Given a set of images $\{x_1, \ldots, x_l, x_{l+1}, \ldots, x_{l+u}\}$ where $n = |V| = l + u$ is the total number of images (i.e. vertices) in the hypergraph $G=(V,E)$. The method constructing the incidence matrix $H$ from the image dataset will be specified clearly in the Experiments and Results section. Let *C* be the number of classes. Please note that $\{x_1, \ldots, x_l\}$ is the set of all labeled points and $\{x_{l+1}, \ldots, x_{l+u}\}$ is the set of all un-labeled points. Let $Y \in R^{|V|*C}$ the initial label matrix for *n* images in the hypergraph *G* be defined as follows:

$$Y_{ij} = \begin{cases} 1 & if\ x_i \in\ class\ j\ and\ 1 \leq i \leq l \\ -1 & if\ x_i \notin\ class\ j\ and\ 1 \leq i \leq l \\ 0 & if\ l+1 \leq i \leq n \end{cases} \tag{6}$$

Let the matrix $F \in R^{|V|*C}$ be the estimated label matrix for the set of images $\{x_1, \ldots, x_l, x_{l+1}, \ldots, x_{l+u}\}$, where the point $x_i$ is labeled as sign $(F_{ij})$ for each class $j$ $(1 \leq j \leq C)$. Our objective is to predict the labels of the un-labeled points $x_{l+1}, \ldots, x_{l+u}$. In the other words, we need to compute the final solution matrix *F*. From [7, 8], the closed form solution of the hypergraph based semi-supervised learning method can be computed as follows:

$$F = (1-\alpha)(I - \alpha D_v^{-\frac{1}{2}} H W D_e^{-1} H^T D_v^{-\frac{1}{2}})^{-1} Y \tag{7}$$

where α is the parameter.

### 3.4. Hypergraph neural network

From [5-6], let:

$$R_1 = ReLU\left(D_v^{-\frac{1}{2}} H W D_e^{-1} H^T D_v^{-\frac{1}{2}} X \theta_1\right) \quad (8)$$

$$R_2 = ReLU(D_v^{-1} H W D_e^{-1} H^T X \theta_1) \quad (9)$$

The output of the hypergraph neural network can be defined and computed as follows:

$$Z = softmax\left(D_v^{-\frac{1}{2}} H W D_e^{-1} H^T D_v^{-\frac{1}{2}} R_1 \theta_2\right) \quad (10)$$

or

$$Z = Z = softmax(D_v^{-1} W D_e^{-1} H^T R_2 \theta_2) \quad (11)$$

Note that $X \in R^{n*L_1}$ is the feature matrix (i.e. the image dataset). $\theta^1 \in R^{L_1*L_2}$ and $\theta^2 \in R^{L_2*C}$ are two parameter matrices that are needed to be learned during the training process.

### 3.5. Our proposed hypergraph neural network

Inspired by the work in [6] for graph neural network, we try to apply this work (idea) to develop our novel version for hypergraph neural network method. In this novel version, we initially compute the final solution matrix $F$ of the classic hypergraph based semi-supervised learning method as in (7):

$$F = (1 - \alpha)\left(I - \alpha D_v^{-\frac{1}{2}} H W D_e^{-1} H^T D_v^{-\frac{1}{2}}\right)^{-1} X \quad (12)$$

In this first step, we overcome the difficulty which is "only neighbors in the two-hop neighborhood are considered" of the graph neural network method and hypergraph neural network method [5-6]. The time complexity of this computation is still low since we employ the sparse matrix computation techniques, for examples, the conjugate gradient method, to solve the above sparse linear system of equations. Please note that, this first step can also be called the "feature propagation" step or "feature diffusion" step [24] (i.e. this step is not a "label propagation" step). Its purpose is to smooth the data (i.e. the feature matrix). This might lead to the high performance of our proposed hypergraph neural network method. In addition, let us define:

$$R_3 = ReLU\left(D_v^{-\frac{1}{2}} H W D_e^{-1} H^T D_v^{-\frac{1}{2}} F \theta_1\right) \quad (13)$$

We compute the final output Z of the hypergraph neural network by:

$$Z = softmax\left(D_v^{-\frac{1}{2}} H W D_e^{-1} H^T R_3 \theta_2\right) \quad (14)$$

Simply speaking, our novel method is the combination of the classic hypergraph based semi-supervised learning method and the hypergraph neural network method. This combination might outperform the hyper- graph neural network alone (i.e. the current state of the art method of semi-supervised learning approach). The two main differences of our proposed method with the method proposed in [6] are:
a) Initially, the input feature matrix X does not need to go through a neural network.
b) At our final step, we compute the output Z of hypergraph neural network similar to the formula proposed by [5]. Please note that [5] proposed the formula to compute the output of the graph neural network. In the other words, in the method proposed in [6], the final solution matrix F of the hypergraph based semi-supervised learning method is just needed to go through only one layer of the neural network which is the softmax layer.

## 4. EXPERIMENTAL RESULT

In this section, we will apply the classic graph based semi-supervised learning method [25], the classic hypergraph based semi-supervised learning method, graph neural network method, hypergraph neural network method, and our proposed hypergraph neural network method to solve the noisy label learning problem. In the other words, we will test the noise robustness of these five methods. The three image datasets that we will use in the experiments are the MNIST dataset, the USPS dataset, and the FASION MNIST dataset.

### 4.1. Dataset

MNIST: This image dataset is the dataset containing the handwritten images from '0' to '9'. There are 70,000 images in the dataset. There are 60,000 images in the training set and 10,000 images in the testing set. Obviously, the number of classes in this MNIST image dataset is 10. Each image in the dataset is the 28-by-28 matrix (gray scale image). Our first task in the preprocessing step is to convert this gray scale image to 1-by-784 vector. We achieve this task by concatenating every rows of the gray scale image to a "long" row vector. In the other words, we have the $R^{70,000 \times 784}$ feature matrix.

USPS: This image dataset is also the handwritten image dataset from '0' to '9'. However, in this dataset, there are just 9,298 images in the dataset. There are 7,291 images in the training set and 2,007 images in the testing set. The number of classes in this USPS dataset is 10. Each image in the dataset is the 16-by-16 matrix (gray scale image). We concatenate every rows of the gray scale image (in this USPS dataset) to the 1-by-256 "long" row vector. Thus, finally, we have the $R^{9,298 \times 256}$ feature matrix.

FASHION MNIST: This image dataset is the dataset containing images of shoes, clothes, caps, etc. There are 70,000 images in the dataset. There 60,000 images in the training set and 10,000 images in the testing set. The number of classes in this FASHION MNIST image dataset is 10. Each image in the dataset is the 28-by-28 matrix (gray scale image). We concatenate every rows of the gray scale image to the 1-by-784 "long" row vector. In the other words, we have the $R^{70,000 \times 784}$ feature matrix. This FASHION MNIST image dataset is considered the hardest image dataset to test in our experiments.

### 4.2. Experiments and result

In order to reduce the noise and redundant features in the input feature matrices and in order to reduce the time constructing the graphs and hypergraphs from the three image datasets, we apply the dimensional reduction PCA technique to the three input feature matrices. Finally, the MNIST dataset is transformed to the $R^{70,000 \times 50}$ matrix. The USPS dataset is transformed to $R^{9,298 \times 50}$ matrix. The FASHION MNIST dataset is transformed to $R^{70,000 \times 300}$ matrix. The way constructing the graphs from the three image datasets can be found in [25, 26]. Next, we will discuss how to construct the incidence matrix $H$ of the hypergraphs from the three image datasets. Please note that the number of hyperedges in the hypergraph is equal to the number of images in the dataset [27]. The image $i$ belongs to hyperedge j if image i is among the k-nearest neighbor of image $j$ or image $j$ is among the k-nearest neighbor of image $i$. In this paper, $k$ is chosen to be 5.

Finally, from the computed H, we can compute the two terms $\left(D_v^{\frac{-1}{2}} H W D_e^{-1} H^T D_v^{\frac{-1}{2}}\right)$ and $(D_v^{-1} H W D_e^{-1} H^T)$ in the symmetric normalized hypergraph Laplacian and the random walk hypergraph Laplacian used in the classic hypergraph based semi-supervised learning method, the hypergraph neural network method, and our proposed hypergraph neural network method. The two main differences of our methods with other semi-supervised learning methods [28, 29] solv- ing the image classification problem with noisy labels are:

a) Other semi-supervised learning methods just used the subsets of the MNIST and the USPS datasets. For example, in [28], the authors just used 10,000 images from MNIST to evaluate their methods. In the other hand, our methods use the complete MNIST, USPS, and FASHION MNIST image datasets.
b) Our methods apply directly the PCA technique to the feature matrices of the three image datasets in order to reduce the time constructing the graphs and the hypergraphs of the three image datasets. To the best of our knowledge, this work has not been done before. The experimental results show that if we do not apply the PCA technique to the feature matrices of the three image dataset, the hypergraph neural network significantly outperforms the graph neural network. If we apply the PCA technique to the feature matrices of the three image datasets, the hypergraph neural network method does not significantly outperform the graph neural network method; however, both the hypergraph neural network and the graph neural network with PCA are better than the graph and hypergraph neural network without using PCA technique. These claims will be clarified in Tables 1, 2, 3, 4.

In general, in this paper, what we want to achieve is clear: we would like to prove that the hypergraph neural networks (the current state of the art semi-supervised learning method and our proposed method) are at least as good as the graph neural network but sometimes lead to better accuracy performance measures. We run our five methods (Python code) on Google Colab with NVIDIA Tesla K80 GPU and 12 GB RAM. The following Tables 1, 2, 3, and 4 show the experimental results of our five methods.

From the above three tables, we easily recognize initially and directly that when the noise level increases, our proposed hypergraph neural network outperforms the other methods (i.e. especially when the noise level reaches 45%). Second, from the experimental results, we see that the hypergraph neural network methods (both the current state of the art semi-supervised learning method and our proposed method) are at least as good as the graph neural network proposed by Thomas Kipf but sometimes lead to better accuracy performance measures. Finally, we can also easily see that the classic graph based semi-supervised learning method performs worst when the noise level increases. Last but not least, in the FASHION MNIST dataset, we would like to show that if we do not apply the PCA technique to the feature matrix of the FASHION MNIST image dataset, the hypergraph neural network significantly outperforms the graph neural network. This claim is shown in the following Table 4.

Table 1. MNIST dataset: comparison of our five methods with various noise levels. The classification accuracy is reported (%)

| Noise level | 0% | 15% | 30% | 45% |
|---|---|---|---|---|
| Graph based semi-supervised learn-ing | 97.70 | 93.33 | 80.94 | 57.4 |
| Hypergraph based semi-supervised learning | 97.65 | 97.56 | 97.54 | 84.49 |
| Graph neural network | 97.39 | 97.31 | 97.11 | 86.15 |
| Hypergraph neural network (current state of the art semi-supervised 97.52 learning method) | 97.52 | 97.40 | **97.34** | 87.07 |
| Proposed hypergraph neural net-work | **97.72** | **97.69** | 97.30 | **91.65** |

Table 2. USPS dataset: comparison of our five methods with various noise levels. The classification accuracy is reported (%)

| Noise level | 0% | 15% | 30% | 45% |
|---|---|---|---|---|
| Graph based semi-supervised learning | **95.06** | **94.96** | 92.82 | 66.26 |
| Hypergraph based semi-supervised learning | 95.06 | 94.91 | **94.71** | 74.58 |
| Graph neural network | 94.66 | 94.42 | 93.12 | 70.00 |
| Hypergraph neural network (current state of the art semi-supervised 97.52 learning method) | 94.76 | 94.71 | 93.82 | 74.48 |
| Proposed hypergraph neural network | **95.06** | 94.81 | 94.37 | **82.51** |

Table 3. Fashion MNIST dataset: comparison of our five methods with various noise levels. The classification accuracy is reported (%)

| Noise level | 0% | 15% | 30% | 45% |
|---|---|---|---|---|
| Graph based semi-supervised learning | 86.13 | 85.75 | 84.02 | 63.61 |
| Hypergraph based semi-supervised learning | 84.88 | 84.69 | 83.43 | 69.29 |
| Graph neural network | **86.76** | **86.35** | 85.17 | 67.79 |
| Hypergraph neural network (current state of the art semi-supervised 97.52 learning method) | 86.41 | 86.30 | **85.47** | 70.09 |
| Proposed hypergraph neural network | 86.14 | 85.91 | 85.02 | **75.89** |

Table 4. Fashion MNIST dataset: Comparison of the hypergraph neural network method (the current state of art semi-supervised learning method) and the graph neural network method

| Noise level | 0% |
|---|---|
| Graph neural network (without PCA) | 79.98 |
| Hypergraph neural network (without PCA) | 85.09 |
| Graph neural network (with PCA) | **86.76** |
| Hypergraph neural network (with PCA) | 86.41 |

## 5. CONCLUSION

In this paper, we have proposed the novel hypergraph neural network method. Our contributions are: (1) Reduce the time constructing the graph and the hypergraph by initially applying the PCA technique to the image dataset, (2) Our novel hypergraph neural network method is in fact the combination of the classic hypergraph based semi-supervised learning method and the hypergraph neural network proposed by [4] (i.e., the current state of the art semi-supervised learning method). The experimental results show that our proposed hypergraph neural network outperforms other semi- supervised learning methods as the noise level in the training set increases (~45%). In the other words, our proposed approach is quite robust to noise labels to some extent. Moreover, the hypergraph neural networks (the current state of the art method of semi-supervised learning approach and our proposed method) are at least as good as the graph neural network proposed by Thomas Kipf, but sometimes lead to better accuracies. Last but not least, in the future work, we will combine the hypergraph p-Laplacian based semi- supervised learning method with the current state of the art semi-supervised learning method (i.e. the hy- pergraph neural network proposed by [4]) to form a novel hypergraph neural network method. Finally, we can apply this novel method to various classification tasks such as protein function prediction, cancer classification, and speech recognition, to name a few.

**ACKNOWLEDGEMENTS**

This research is funded by Ho Chi Minh City University of Technology (HCMUT), VNU-HCM, under grant number BK-SDH-2020-1970428.

**REFERENCES**
[1]  Karen Simonyan and Andrew Zisserman, "Very deep convolutional networks for large-scale image recognition," a*rXiv preprint arXiv:1409.1556,* pp. 1-14, 2015.
[2]  G. Hinton *et al*., "Deep Neural Networks for Acoustic Modeling in Speech Recognition: The Shared Views of Four Research Groups," in *IEEE Signal Processing Magazine*, vol. 29, no. 6, pp. 82-97, 2012, doi: 10.1109/MSP.2012.2205597.
[3]  Thomas N. Kipf, Max Welling, "Semi-supervised classification with graph convolutional net-works," *arXiv preprint arXiv:1609.02907*, pp. 1-14, 2016.
[4]  Yifan Feng, Haoxuan You, Zizhao Zhang, Rongrong Ji and Yue Gao, "Hypergraph neural networks," *The Thirty-Third AAAI Conference on Artificial Intelligence (AAAI-19)*, Honolulu, vol. 33, no. 1, pp. 3558-3565, 2019, DOI: https://doi.org/10.1609/aaai.v33i01.33013558.
[5]  Song Bai, Feihu Zhang, Philip H.S. Torr, "Hypergraph convolution and hypergraph attention," *arXiv preprint arXiv:1901.08150,* pp. 1-30, 2020.
[6]  Klicpera, Johannes, Aleksandar Bojchevski, and Stephan Gu¨nnemann, "Predict then Propagate: Graph Neural Networks meet Personalized PageRank," *arXiv preprint arXiv:1810.05997,* vol. 1, 2018.
[7]  [8] Dengyong Zhou, Jiayuan Huang and Bernhard Scholkopf, "Learning with Hypergraphs: Clustering, Classification, and Embedding," in *Advances in Neural Information Processing Systems 19: Proceedings of the 2006 Conference*, MIT Press, pp. 1601-1608, 2006.
[8]  Dengyong Zhou, Jiayuan Huang and Bernhard Scholkopf, "Beyond pairwise classification and clustering using hypergraphs (Technical Report 143)." *Max Plank Institute for Biological Cybernetics*, Tübingen, Germany*,* 2005.
[9]  Zijin Zhao, "Classification in the presence of heavy label noise: A Markov chain sampling framework". *Theses (School of Computing Science), Simon Fraser University*, 2017.
[10] K. Yi and J. Wu, "Probabilistic End-To-End Noise Correction for Learning With Noisy Labels," *2019 IEEE/CVF Conference on Computer Vision and Pattern Recognition (CVPR)*, Long Beach, CA, USA, pp. 7010-7018, 2019, doi: 10.1109/CVPR.2019.00718.
[11] Chiyuan Zhang, Samy Bengio, Moritz Hardt, Benjamin Recht, Oriol Vinyals, "Understanding deep learning requires rethinking generalization," *arXiv preprint arXiv:1611.03530*, pp. 1-15, 2017.
[12] D. Tanaka, D. Ikami, T. Yamasaki and K. Aizawa, "Joint Optimization Framework for Learning with Noisy Labels," *2018 IEEE/CVF Conference on Computer Vision and Pattern Recognition*, Salt Lake City, UT, pp. 5552-5560, 2018, doi: 10.1109/CVPR.2018.00582.
[13] Mikael Henaff, Joan Bruna and Yann LeCun, "Deep convolutional networks on graph-structured data," *arXiv preprint arXiv:1506.05163*, pp. 1-10, 2015.
[14] Jakramate Bootkrajang and Ata Kabán, "Label-noise robust logistic regression and its applications," in *Machine Learning and Knowledge Discovery in Databases: proceeding of Joint European Conference on Machine Learning and Knowledge Discovery in Databases*, Springer, pp. 143-158, 2012.
[15] Jakramate Bootkrajang, "Supervised learning with random labelling errors," *Computer Science, University of Birmingham,* 2013.
[16] A. J. Bekker and J. Goldberger, "Training deep neural-networks based on unreliable labels," *2016 IEEE International Conference on Acoustics, Speech and Signal Processing (ICASSP)*, Shanghai, pp. 2682-2686, 2016, doi: 10.1109/ICASSP.2016.7472164.
[17] Jacob Goldberger and Ehud Ben-Reuven "Training deep neural-networks using a noise adaptation layer," *5th International Conference on Learning Representations*, Toulon, pp. 1-7, 2017.
[18] A. J. Bekker, M. Chorev, L. Carmel and J. Goldberger, "A deep neural network witharestricted noisy channel for identification of functional introns," *2017 IEEE 27th International Workshop on Machine Learning for Signal Processing (MLSP)*, Tokyo, pp. 1-6, 2017, doi: 10.1109/MLSP.2017.8168186.
[19] Aritra Ghosh, Himanshu Kumar and P.S. Sastry, "Robust loss functions under label noise for deep neural networks," *Thirty-First AAAI Conference on Artificial Intelligence*, San Francisco, pp. 1919-1925, 2017.
[20] Zhilu Zhang and Mert R. Sabuncu, "Generalized cross entropy loss for training deep neural networks with noisy labels," *NIPS'18: Proceedings of the 32nd International Conference on Neural Information Processing Systems*, New York, pp. 8792-8802, 2018.
[21] B. Frenay and M. Verleysen, "Classification in the Presence of Label Noise: A Survey," in *IEEE Transactions on Neural Networks and Learning Systems*, vol. 25, no. 5, pp. 845-869, 2014, doi: 10.1109/TNNLS.2013.2292894.
[22] Wei Shen, Kai Zhao, Yilu Guo and Alan Yuille, "Label distribution learning forests," *NIPS'17: Proceedings of the 31st International Conference on Neural Information Processing Systems*, New York, pp. 834-843, 2017.

[23] X. Geng, "Label Distribution Learning," in *IEEE Transactions on Knowledge and Data Engineering*, vol. 28, no. 7, pp. 1734-1748, 1 July 2016, doi: 10.1109/TKDE.2016.2545658.
[24] Ryan A. Rossi, Rong Zhou and Nesreen K. Ahmed "Deep feature learning for graphs," *arXiv preprint arXiv:1704.08829,* pp. 1-11, 2017.
[25] Dengyong Zhou. Olivier Bousquet, Thomas Navin Lal, Jason Weston and Bernhard H Schölkopf, "Learning with local and global consistency," *NIPS'03: Proceedings of the 16th International Conference on Neural Information Processing Systems*, vol. 16, no. 3, pp. 321-328, 2004.
[26] Ulrike von Luxburg, "A tutorial on spectral clustering," *Statistics and computing*, vol. 17, pp. 395-416, 2007, https://doi.org/10.1007/s11222-007-9033-z.
[27] Y. Huang, Q. Liu, S. Zhang and D. N. Metaxas, "Image retrieval via probabilistic hypergraph ranking," *2010 IEEE Computer Society Conference on Computer Vision and Pattern Recognition*, San Francisco, CA, pp. 3376-3383, 2010, doi: 10.1109/CVPR.2010.5540012.
[28] B. Jiang, Z. Zhang, D. Lin, J. Tang and B. Luo, "Semi-Supervised Learning With Graph Learning-Convolutional Networks," *2019 IEEE/CVF Conference on Computer Vision and Pattern Recognition (CVPR)*, Long Beach, CA, USA, pp. 11305-11312, 2019, doi: 10.1109/CVPR.2019.01157.
[29] Zhiwu Lu and Liwei Wang, "Noise-robust semi-supervised learning via fast sparse coding," *Pattern Recognition* vol. 48, no. 2, pp. 605-612, 2015.